\documentclass{article}
\usepackage{graphicx} 

\usepackage{arxiv}

\usepackage{amsmath}
\usepackage[utf8]{inputenc} 
\usepackage[T1]{fontenc}    
\usepackage{hyperref}       
\usepackage{url}            
\usepackage{booktabs}       
\usepackage{amsfonts}       
\usepackage{nicefrac}       
\usepackage{microtype}      
\usepackage[numbers]{natbib}
\usepackage{doi}
\usepackage{caption}

\title{On the Computation of the Fisher Information in Continual Learning}
\date{April, 2025}

\author{ \href{https://orcid.org/0000-0000-0000-0000}{\includegraphics[scale=0.06]{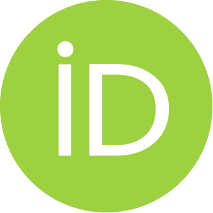}\hspace{1mm}Gido M.~van de Ven} \\
	KU Leuven (Belgium), TU Delft (the Netherlands)\\
	\texttt{gido.vandeven@kuleuven.be} \\
}

\renewcommand{\headeright}{ICLR Blogpost}
\renewcommand{\undertitle}{ICLR Blogpost}
\renewcommand{\shorttitle}{On the Computation of the Fisher Information}

\hypersetup{
pdftitle={On the Computation of the Fisher Information in Continual Learning},
pdfsubject={cs.LG, cs.AI, stat.ML, cs.CV},
pdfauthor={Gido M.~van de Ven},
pdfkeywords={Continual learning, Elastic Weight Consolidation, Fisher Information},
}

\begin{document}

\maketitle

\begin{abstract}
	One of the most popular methods for continual learning with deep neural networks is Elastic Weight Consolidation (EWC), which involves computing the Fisher Information. The exact way in which the Fisher Information is computed is however rarely described, and multiple different implementations for it can be found online. This blog post discusses and empirically compares several often-used implementations, which highlights that many currently reported results for EWC could likely be improved by changing the way the Fisher Information is computed.
\end{abstract}


\section{Introduction}
Continual learning is a rapidly growing subfield of deep learning devoted to enabling neural networks to incrementally learn new tasks, domains or classes while not forgetting previously learned ones. Such continual learning is crucial for addressing real-world problems where data are constantly changing, such as in healthcare, autonomous driving or robotics. Unfortunately, continual learning is challenging for deep neural networks, mainly due to their tendency to forget previously acquired skills when learning something new.

Elastic Weight Consolidation (EWC)~\cite{kirkpatrick2017overcoming}, developed by Kirkpatrick and colleagues from DeepMind, is one of the most popular methods for continual learning with deep neural networks. To this day, this method is featured as a baseline in a large proportion of continual learning studies. However, in the original paper the exact implementation of EWC was not well described, and no official code was provided. A previous blog post by Huszár~\cite{huszar2017comment} already addressed an issue relating to how EWC should behave when there are more than two tasks.\footnote{In this blog post, I use the ``online'' version of EWC described by Huszár~\cite{huszar2017comment}.} This blog post deals with the question of how to compute the Fisher Information matrix. The Fisher Information plays a central role in EWC, but the original paper does not detail how it should be computed. Other papers using EWC also rarely describe how they compute the Fisher Information, even though various different implementations for doing so can be found online.

The Fisher Information matrix is also frequently used in the optimization literature. In this literature, several years ago, Kunstner and colleagues~\cite{kunstner2019limitations} discussed two ways of computing the Fisher Information --- the `true' Fisher and the `empirical' Fisher --- and based on both theory and experiments they recommended against using the empirical Fisher approximation. It seems however that this discussion has not reached the continual learning community. In fact, as we will see, the most commonly used way of computing the Fisher Information in continual learning makes even cruder approximations than the empirical Fisher.

\section{The Continual Learning Problem}

Before diving into EWC and the computation of the Fisher Information, let me introduce the continual learning problem by means of a simple example. Say, we have a deep neural network model $f_{\boldsymbol{\theta}}$, parameterized by weight vector $\boldsymbol{\theta}$. This model has already been trained on a first task (or a first set of tasks, this example can work recursively), by optimizing a loss function $\ell_{\text{old}}(\boldsymbol{\theta})$ on training data $D_{\text{old}}\sim\mathcal{D}_{\text{old}}$. This resulted in weights $\hat{\boldsymbol{\theta}}_{\text{old}}$. We then wish to continue training this model on a new task, by optimizing a loss function $\ell_{\text{new}}(\boldsymbol{\theta})$ on training data $D_{\text{new}}\sim\mathcal{D}_{\text{new}}$, in such a way that the model maintains, or possibly improves, its performance on the previously learned task(s). Unfortunately, as has been thoroughly described in the continual learning literature, if the model is continued to be trained on the new data in the standard way (i.e., optimizing $\ell_{\text{new}}(\boldsymbol{\theta})$ with stochastic gradient descent), the typical result is \emph{catastrophic forgetting}~\cite{mccloskey1989catastrophic,ratcliff1990connectionist}: a model that is good for the new task, but no longer for the old one(s).

In this blog post, similar to most of the deep learning work on continual learning, the focus is on supervised learning. Each data point thus consists of an input $\boldsymbol{x}$ and a corresponding output $y$, and our deep neural network models the conditional distribution $p_{\boldsymbol{\theta}}(y|\boldsymbol{x})$.

\section{Elastic Weight Consolidation}

Now we are ready to take a detailed look at EWC. We start by formally defining this method. When training on a new task, to prevent catastrophic forgetting, rather than optimizing only the loss on the new task $\ell_{\text{new}}(\boldsymbol{\theta})$, EWC adds an extra term to the loss that involves the Fisher Information:

$$
\ell_{\text{EWC}}(\boldsymbol{\theta})=
\ell_{\text{new}}(\boldsymbol{\theta})+\frac{\lambda}{2}\sum_{i=1}^{N_{\text{params}}}
F_{\text{old}}^{i,i}(\theta^{i}-\hat{\theta}_{\text{old}}^{i})^2
$$

In this expression, $N_{\text{params}}$ is the number of parameters in the model, $\theta^{i}$ is the value of parameter $i$ (i.e., the $i^{\text{th}}$ element of weight vector $\boldsymbol{\theta}$), $F_{\text{old}}^{i,i}$ is the $i^{\text{th}}$ diagonal element of the model's Fisher Information matrix on the old data, and $\lambda$ is a hyperparameter that sets the relative importance of the new task compared to the old one(s).

EWC can be motivated from two perspectives, each of which I discuss next.

\subsection{Penalizing Important Synapses}
Loosely inspired by neuroscience theories of how synapses in the brain critical for previously learned skills are protected from overwriting during subsequent learning~\cite{yang2009stably}, a first motivation for EWC is that when training on a new task, large changes to network parameters  important for previously learned task(s) should be avoided. To achieve this, for each parameter $\theta^{i}$, the term $F_{\text{old}}^{i,i}(\theta^{i}-\hat{\theta}_{\text{old}}^{i})^2$ penalizes changes away from $\hat{\theta}_{\text{old}}^{i}$, which was that parameter's optimal value after training on the old data. Importantly, how strongly these changes are penalized differs between parameters. This strength is set by $F_{\text{old}}^{i,i}$, the $i^{\text{th}}$ diagonal element of the network's Fisher Information matrix on the old data, which is used as a proxy for how important that parameter is for the old tasks. The diagonal elements of the Fisher are a sensible choice for this, as they measure how much the network's output would change due to small changes in each of its parameters.

\subsection{Bayesian Perspective}
A second motivation for EWC comes from a Bayesian perspective, because EWC can also be interpreted as performing approximate Bayesian inference on the parameters of the neural network. For this we need to take a probabilistic perspective, meaning that we view the network parameters $\boldsymbol{\theta}$ as a random variable over which we want to learn a distribution. Then, when learning a new task, the idea behind EWC is to use the posterior distribution $p(\boldsymbol{\theta}|D_{\text{old}})$ that was found after training on the old task(s), as the prior distribution when training on the new task. To make this procedure tractable, the Laplace approximation is used, meaning that the distribution $p(\boldsymbol{\theta}|D_{\text{old}})$ is approximated as a Gaussian centered around $\hat{\boldsymbol{\theta}}_{\text{old}}$ and with the Fisher Information $F_{\text{old}}$ as precision matrix. To avoid letting the computational costs become too high, EWC sets the diagonal elements of $F_{\text{old}}$ to zero.\footnote{See~\cite{ritter2018online} for an extension of EWC that relaxes this simplification.} For a more in-depth treatment of EWC from a Bayesian perspective, I refer to~\cite{huszar2017comment,aich2021elastic}.

\section{A Closer Look at the Fisher Information}

EWC thus involves computing the diagonal elements of the network's Fisher Information on the old data. Following the definitions and notation in Martens~\cite{martens2020new}, the $i^{\text{th}}$ diagonal element of this Fisher Information matrix is defined as:

\begin{equation}
F_{\text{old}}^{i,i} := \mathbb{E}_{\boldsymbol{x}\sim\mathcal{D}_{\text{old}}} \left[ \ \mathbb{E}_{y\sim p_{\hat{\boldsymbol{\theta}}_{\text{old}}}} \left[ \left(  \left. \frac{\delta \log{p_{\boldsymbol{\theta}}\left(y|\boldsymbol{x}\right)}}{\delta \theta^i}\right\rvert_{\boldsymbol{\theta}=\hat{\boldsymbol{\theta}}_{\text{old}}} \right)^2 \right] \right]
\label{eq:fisher}
\end{equation}

In this definition, there are two expectations: (1) an outer expectation over $\mathcal{D}_{\text{old}}$, which is the (theoretical) input distribution of the old data; and (2) an inner expectation over $p_{\hat{\boldsymbol{\theta}}_{\text{old}}}(y|\boldsymbol{x})$, which is the conditional distribution of $y$ given $\boldsymbol{x}$ defined by the neural network after training on the old data. The different ways of computing the Fisher Information that can be found in the continual learning literature differ in how these two expectations are computed or approximated.

\section{Different Ways of Computing the Fisher Information}

\subsection{Exact}
If computational costs are not an issue, the outer expectation in Eq~(\ref{eq:fisher}) can be estimated by averaging over all available training data $D_{\text{old}}$, while --- in the case of a classification problem --- the inner expectation can be calculated for each training sample exactly:

$$
F_{\text{old, EXACT}}^{i,i} = \frac{1}{|D_{\text{old}}|} \sum_{\boldsymbol{x}\in D_{\text{old}}} \left( 
\sum_{y=1}^{N_{\text{classes}}} p_{\hat{\boldsymbol{\theta}}_{\text{old}}}\left(y|\boldsymbol{x}\right) \left( \left. 
\frac{\delta\log p_{\boldsymbol{\theta}}\left(y|\boldsymbol{x}\right)}{\delta\theta^i} 
\right\rvert_{\boldsymbol{\theta}=\hat{\boldsymbol{\theta}}_{\text{old}}} \right)^2 \right)
$$

I refer to this option as \textbf{EXACT}, because for each sample in $D_{\text{old}}$, the diagonal elements of the Fisher Information are computed exactly. I am not aware of many implementations of EWC that use this way of computing the Fisher Information, but one example can be found in~\cite{van2022three}. A disadvantage of this option is that it can be computationally costly, especially if the number of training samples and/or the number of possible classes is large, because for each training sample a separate gradient must be computed for every possible class.

\subsection{Sampling Data Points}
One way to reduce the costs of computing $F^{i,i}_{\text{old}}$ is 
by estimating the outer expectation using only a subset of the old training data:

$$
F_{\text{old, EXACT}(n)}^{i,i} = \frac{1}{n} \sum_{\boldsymbol{x}\in S_{D_{\text{old}}}^{(n)}} \left( \sum_{y=1}^{N_{\text{classes}}} p_{\hat{\boldsymbol{\theta}}_{\text{old}}}\left(y|\boldsymbol{x}\right) \left( \left. \frac{\delta\log p_{\boldsymbol{\theta}}\left(y|\boldsymbol{x}\right)}{\delta\theta^i}\right\rvert_{\boldsymbol{\theta}=\hat{\boldsymbol{\theta}}_{\text{old}}} \right)^2 \right)
$$

whereby $S_{D_{\text{old}}}^{(n)}$ is a set of $n$ random samples from $D_{\text{old}}$. Although this seems a natural way to reduce the computational costs of computing the Fisher Information, I am aware of only one study~\cite{benzing2022unifying} that has implemented EWC in this way. Below, we will explore EWC with this implementation using $n=500$. I refer to this option as \textbf{EXACT (\emph{n}=500)}, because for each data point that is considered, it is still the case that the exact version of the Fisher's diagonal elements are computed.

\subsection{Sampling Labels}
Another way to make the computation of $F^{i,i}_{\text{old}}$ less costly is by computing the squared gradient not for all possible classes, but only for a single class per training sample. This means that the inner expectation in the definition of $F^{i,i}_{\text{old}}$ is no longer computed exactly. To maintain an unbiased estimate of the inner expectation, Monte Carlo sampling can be used. That is, for each given training sample $\boldsymbol{x}$, the class for which to compute the squared gradient can be selected by sampling from $p_{\hat{\boldsymbol{\theta}}_{\text{old}}}\left(.|\boldsymbol{x}\right)$. This gives:

$$
F_{\text{old, SAMPLE}}^{i,i} = \frac{1}{|D_{\text{old}}|} \sum_{\boldsymbol{x}\in D_{\text{old}}} \left( \left. \frac{\delta\log p_{\boldsymbol{\theta}}\left(c_{\boldsymbol{x}}|\boldsymbol{x}\right)}{\delta\theta^i} \right\rvert_{\boldsymbol{\theta}=\hat{\boldsymbol{\theta}}_{\text{old}}} \right)^2
$$

whereby, independently for each $\boldsymbol{x}$, $c_{\boldsymbol{x}}$ is randomly sampled from $p_{\hat{\boldsymbol{\theta}}_{\text{old}}}(.|\boldsymbol{x})$. I refer to this option as \textbf{SAMPLE}. This way of unbiasedly estimating the Fisher Information has been used in the implementation of EWC in~\cite{liu2018rotate,kao2021natural}.

\subsection{Empirical Fisher}
Another option is to compute the squared gradient only for each sample's ground-truth class:

$$
F_{\text{old, EMPIRICAL}}^{i,i} = \frac{1}{|D_{\text{old}}|} \sum_{\left(\boldsymbol{x},y\right)\in D_{\text{old}}} \left( \left. \frac{\delta\log p_{\boldsymbol{\theta}}\left(y|\boldsymbol{x}\right)}{\delta\theta^i} \right\rvert_{\boldsymbol{\theta}=\hat{\boldsymbol{\theta}}_{\text{old}}} \right)^2
$$

Computed this way, $F_{\text{old}}$ corresponds to the  ``empirical'' Fisher Information matrix~\cite{martens2020new}. I therefore refer to this option as \textbf{EMPIRICAL}. Chaudhry and colleagues~\cite{chaudhry2018riemannian} advocated for using this option when implementing EWC. Their argument is that the ``true'' Fisher~(i.e., the option to which in this blog post I refer as EXACT) is computationally too expensive, and that, because at a good optimum the model distribution $p_{\hat{\boldsymbol{\theta}}_{\text{old}}}(.|\boldsymbol{x})$ approaches the ground-truth output distribution, the empirical Fisher is expected to behave in a similar manner as the true Fisher. However, as mentioned in the introduction, in the optimization literature, researchers have cautioned against using the empirical Fisher as approximation of the true Fisher~\cite{kunstner2019limitations}. Nevertheless, in continual learning, it still appears to be rather common to implement EWC using the empirical Fisher, or --- as we will see next --- an approximate version of the empirical Fisher.

\subsection{Batched Approximation of Empirical Fisher}
The last option that we consider has probably come about thanks to a feature of PyTorch. Note that all of the above ways of computing $F^{i,i}_{\text{old}}$ require access to the gradients of the individual data points, as the gradients need to be squared before being summed. However, batch-wise operations in PyTorch only allow access to the aggregated gradients, not to the individual, unaggregated gradients. In PyTorch, the above ways of computing $F^{i,i}_{\text{old}}$ could therefore only be implemented with mini-batches of size one. Perhaps in an attempt to gain efficiency, several implementations of EWC can be found on Github that compute $F^{i,i}_{\text{old}}$ by squaring the aggregated gradients of mini-batches of size larger than one.  Indeed, popular continual learning libraries such as Avalanche\footnote{\url{https://github.com/ContinualAI/avalanche/blob/c1ca18d1c44f7cc8964686efd54a79443763d945/avalanche/training/plugins/ewc.py\#L161-L180}.}~\cite{carta2023avalanche} and PyCIL\footnote{\url{https://github.com/G-U-N/PyCIL/blob/0cb8ad6ca6da93deff5e8767cfb143ed2aa05809/models/ewc.py\#L234-L254}.}~\cite{zhou2023pycil} use this approach, which probably makes this variant of computing the Fisher the one that is most used in the continual learning literature. Typically, these batched implementations only use the gradients for the ground-truth classes (i.e., they are approximate versions of the empirical Fisher):

$$
F_{\text{old, BATCHED}(b)}^{i,i} = \frac{1}{|D_{\text{old}}^{(b)}|} \sum_{\mathcal{B}\in D_{\text{old}}^{(b)}} \left( \sum_{\left(\boldsymbol{x},y\right)\in \mathcal{B}} \left. \frac{\delta\log p_{\boldsymbol{\theta}}\left(y|\boldsymbol{x}\right)}{\delta\theta^i} \right\rvert_{\boldsymbol{\theta}=\hat{\boldsymbol{\theta}}_{\text{old}}} \right)^2
$$

whereby $D_{\text{old}}^{(b)}$ is a batched version of the old training data $D_{\text{old}}$, so that the elements of $D_{\text{old}}^{(b)}$ are mini-batches with $b$ training samples. (And $|D_{\text{old}}^{(b)}|$ is the number of mini-batches, not the number of training samples.) Below, we will explore this option using $b=128$, referring to it as \textbf{BATCHED (\emph{b}=128)}.

\section{Empirical Comparisons}

Now, let us empirically compare the performance of EWC with these various ways of computing the Fisher Information. To do so, I use two relatively simple, often used continual learning benchmarks: Split MNIST and Split CIFAR-10. For these benchmarks, the original MNIST or CIFAR-10 dataset is split up into five tasks with two classes per task. Both benchmarks are performed according to the task-incremental learning scenario, using a separate softmax output layer for each task. For Split MNIST, following~\cite{van2022three}, a fully connected network is used with two hidden layers of 400~ReLUs each. For Split CIFAR-10, following~\cite{lopez2017gradient,hess2024two}, a reduced ResNet-18 is used without pre-training. For both benchmarks, the Adam-optimizer~\cite{kingma2015adam} ($\beta_1$=0.9, $\beta_2$=0.999) is used to train for 2000 iterations per task with stepsize of 0.001 and mini-batch size of 128 (Split MNIST) or 256 (Split CIFAR). Each experiment is run 30 times with different random seeds, and reported are the mean $\pm$ standard error over these runs. Code to replicate these experiments is available at \url{https://github.com/GMvandeVen/continual-learning}.

\subsection{Split MNIST}
For the experiments on Split MNIST, the results are shown in Figure~\ref{fig:mnist} and Table~\ref{tab:mnist}.

\begin{table}[ht]
\vskip -0.1in
	\begin{minipage}{0.55\linewidth}
		\centering
		\vskip 0in
		\includegraphics[width=\textwidth]{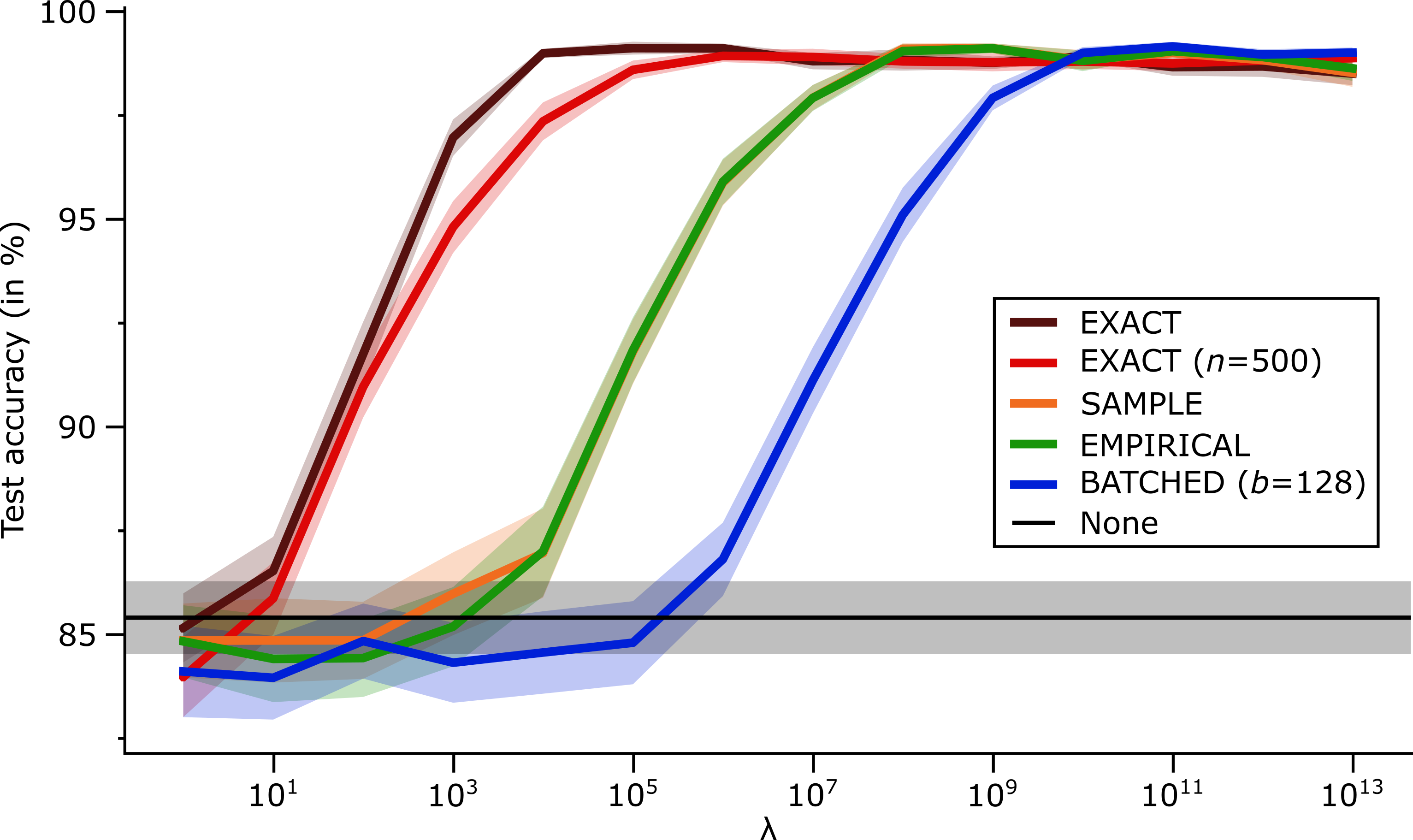}
		\vskip 0.11in
		\captionof{figure}{\label{fig:mnist} \textbf{Split MNIST.} Performance of EWC with different ways of computing the Fisher Information for a wide range of hyperparameter values. 
        }
	\end{minipage}\hfill
	\begin{minipage}{0.42\linewidth}
      \centering
      \vskip 0.42in
      \begin{center}
      \begin{footnotesize}
      \resizebox{\textwidth}{!}{
       \begin{tabular}{lcc} \toprule
          & \textbf{Accuracy} & \textbf{Train time} \\ \midrule
         EXACT & 99.12 ($\pm$ 0.16) & 121 ($\pm$ 1) \\
         EXACT ($n$=500) & 98.93 ($\pm$ 0.15) & ~~58 ($\pm$ 0) \\
         SAMPLE & 99.12 ($\pm$ 0.12) & 101 ($\pm$ 1) \\
         EMPIRICAL & 99.12 ($\pm$ 0.12) & 100 ($\pm$ 1)  \\
         BATCHED ($b$=128) & 99.11 ($\pm$ 0.16) & ~~58 ($\pm$ 1) \\ \midrule
         \emph{None} & \emph{85.41 ($\pm$ 0.88)} & \emph{~~53 ($\pm$ 0)} \\
         \bottomrule
       \end{tabular}
      }
      \vskip 0.39in
      \caption{\label{tab:mnist} \textbf{Split MNIST.} The average final test accuracy (in \%) for the best performing hyperparameter value of each variant, and the total training time (in seconds) on an NVIDIA RTX 2000 Ada Generation GPU. 
      }
      \end{footnotesize}
      \end{center}
      \vskip -0.2in
	\end{minipage}
\end{table}

From Table~\ref{tab:mnist}, we can see that for Split MNIST, when looking only at the performance of the best performing hyperparameter, there are no substantial differences between the various ways of computing the Fisher. However, from Figure~\ref{fig:mnist}, we can see that there are large differences in terms of the range of hyperparameter values that EWC performs well with. For example, when using the BATCHED option of computing the Fisher, EWC requires a hyperparameter orders of magnitude larger than the best hyperparameter for the EXACT option. This suggests that there might be important differences between these different ways of computing the Fisher, but that perhaps the task-incremental version of Split MNIST is not difficult enough to elicit significant differences in the best performance between them.

\subsection{Split CIFAR-10}
Therefore, let us look at the more difficult Split CIFAR-10 benchmark, for which the results are shown in Figure~\ref{fig:cifar} and Table~\ref{tab:cifar}.


\begin{table}[ht]
\vskip -0.1in
	\begin{minipage}{0.55\linewidth}
		\centering
		\vskip 0in
		\includegraphics[width=\textwidth]{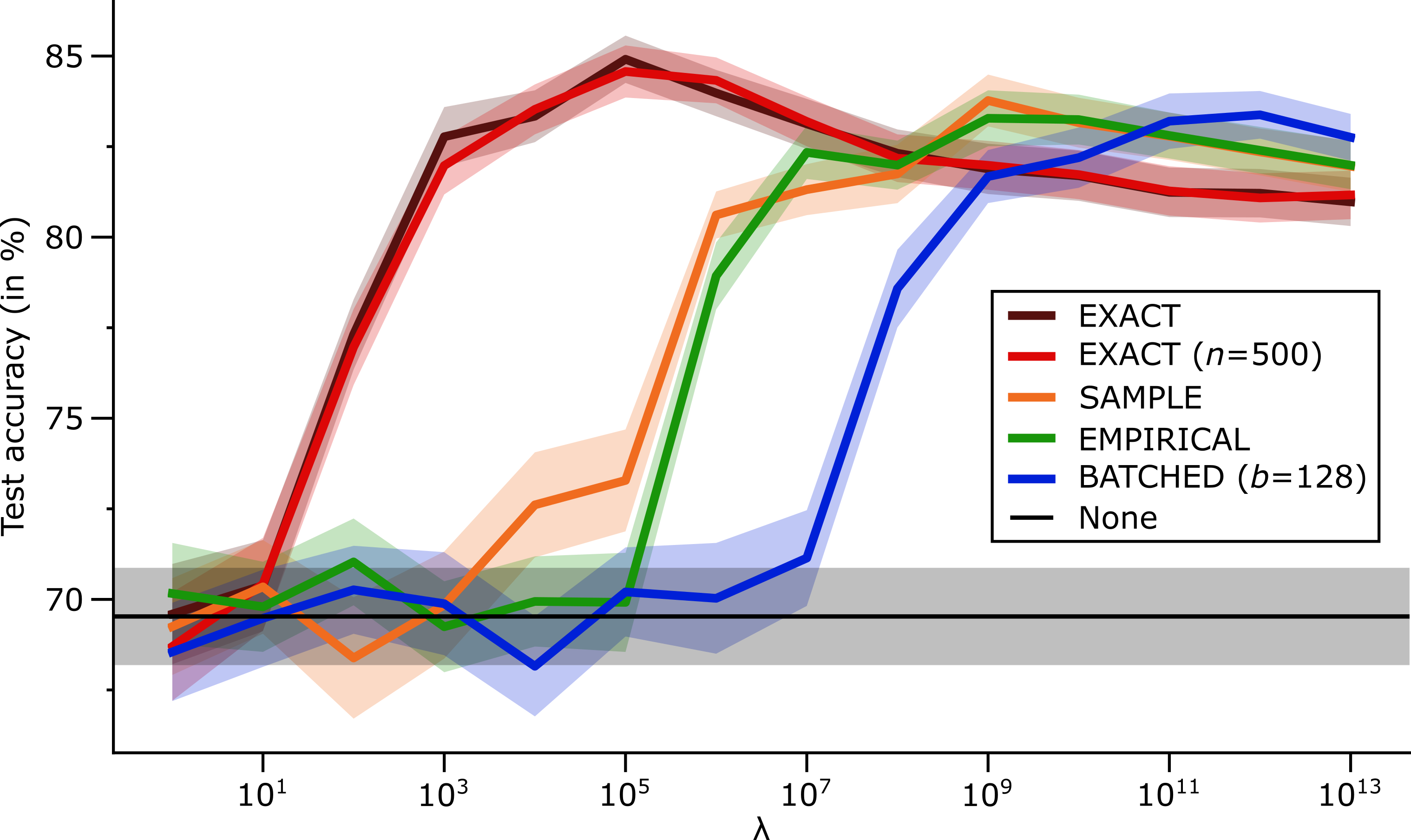}
		\vskip 0.11in
		\captionof{figure}{\label{fig:cifar} \textbf{Split CIFAR-10.} Performance of EWC with different ways of computing the Fisher Information for a wide range of hyperparameter values. 
        }
	\end{minipage}\hfill
	\begin{minipage}{0.42\linewidth}
      \centering
      \vskip 0.42in
      \begin{center}
      \begin{footnotesize}
      \resizebox{\textwidth}{!}{
       \begin{tabular}{lcc} \toprule
          & \textbf{Accuracy} & \textbf{Train time} \\ \midrule
         EXACT & 84.91 ($\pm$ 0.65) & 972 ($\pm$ 5) \\
         EXACT ($n$=500) & 84.57 ($\pm$ 0.72) & 668 ($\pm$ 1) \\
         SAMPLE & 83.77 ($\pm$ 0.72) & 817 ($\pm$ 2) \\
         EMPIRICAL & 83.28 ($\pm$ 0.77) & 817 ($\pm$ 2) \\
         BATCHED ($b$=128) & 83.38 ($\pm$ 0.66) & 667 ($\pm$ 1) \\ \midrule
         \emph{None} & \emph{69.53 ($\pm$ 1.34)} & \emph{627 ($\pm$ 3)} \\
         \bottomrule
       \end{tabular}
      }
      \vskip 0.39in
      \caption{\label{tab:cifar} \textbf{Split CIFAR-10.} The average final test accuracy (in \%) for the best performing hyperparameter value of each variant, and the total training time (in seconds) on an NVIDIA RTX 2000 Ada Generation GPU. 
      }
      \end{footnotesize}
      \end{center}
      \vskip -0.2in
	\end{minipage}
\end{table}



Indeed, on this benchmark, there are significant differences between the different options also in terms of their best performance. The performance of EWC is substantially better when the Fisher Information is computed exactly, even when this is done only for a subset of the old training data, compared to when it is estimated or approximated in same way. We can further see that the SAMPLE option, which uses an unbiased estimate of the true Fisher, appears to perform somewhat better than using the empirical Fisher, but the difference is small and non-conclusive. Interestingly, also on this more difficult benchmark, using the batched approximation of the empirical Fisher still results in a similar best performance as using the regular empirical Fisher, although these two options do differ in terms of their optimal hyperparameter range.

\section{Conclusion and Recommendations}

I finish this blog post by concluding that the way in which the Fisher Information is computed can have a substantial impact on the performance of EWC. This is an important realization for the continual learning research community. Going forwards, based on my findings, I have three recommendations for researchers in this field.
Firstly, whenever using EWC --- or another method that uses the Fisher Information --- make sure to describe the details of how the Fisher Information is computed. Secondly, do not simply ``use the best performing hyperparameter(s) from another paper'', especially if you cannot guarantee that the details of your implementation are the same as in the other paper. And thirdly, when using the Fisher Information matrix, it is preferable to compute it exactly rather than approximating it. If computational resources are scarce, it seems better to reduce the number of training samples used to compute the Fisher, than to cut corners in another way.

\subsection*{Acknowledgements}

This work has been supported by a senior postdoctoral fellowship from the Resarch Foundation -- Flanders (FWO) under grant number 1266823N.

\bibliographystyle{unsrtnat}
\bibliography{main}

\end{document}